\documentclass[11pt,a4paper]{article}

\usepackage[margin=1in]{geometry}
\usepackage{amsmath,amssymb,amsthm}
\usepackage{booktabs}
\usepackage{graphicx}
\usepackage[colorlinks=true,linkcolor=blue,citecolor=blue,urlcolor=blue]{hyperref}
\usepackage{xcolor}
\usepackage{tikz}
\usepackage{pgfplots}
\pgfplotsset{compat=1.18}
\usetikzlibrary{positioning,shapes.geometric,arrows.meta,
                backgrounds,decorations.pathreplacing,calc}
\usepackage[numbers,compress]{natbib}
\usepackage{parskip}
\usepackage{titlesec}
\titlespacing*{\section}{0pt}{1.2ex plus .5ex minus .2ex}{0.8ex plus .2ex}
\titlespacing*{\subsection}{0pt}{1.0ex plus .4ex minus .2ex}{0.6ex plus .2ex}

\newtheorem{observation}{Observation}

\definecolor{detectgreen}{RGB}{56,142,60}
\definecolor{absorbred}{RGB}{198,40,40}
\definecolor{clusterblue}{RGB}{21,101,192}
\definecolor{clusterteal}{RGB}{0,131,143}

\title{\textbf{Squish and Release: Exposing Hidden Hallucinations\\
by Making Them Surface as Safety Signals}}

\author{
  Nathaniel Oh \quad Paul Attie \\
  School of Computer and Cyber Sciences \\
  Augusta University \\
  Augusta, Georgia 30912 \\
  \texttt{\{noh,pattie\}@augusta.edu}
}

\date{}

\begin{document}
\maketitle
\thispagestyle{plain}
\footnotetext{Code, data, and pre-built activation vectors: \url{https://github.com/calysteon/order-gap-benchmark}}

\begin{abstract}
Hallucination under conversational pressure is invisible
to output inspection: the model produces authoritative,
confident professional output built on a false premise
it already knew was wrong, and no examination of that
output reveals the error. The central insight of this
work is that the error is not in the output-it is in
the activation space of the safety circuit, suppressed
by accumulated conversational pressure. Like a signal
masked by noise, the detection is present but not
surfacing. We introduce \emph{Squish and Release}
(S\&R), a perceptual lens architecture that makes
this hidden signal visible. S\&R has two separable
components: a \emph{detector body}-layers 24--31 of
the residual stream, the localized circuit where safety
evaluation lives-and a swappable \emph{detector core},
an activation vector patched into that circuit to
control what the model perceives and flags. The core
is directional: a \emph{safety core} (captured from a
maximally-refused prompt) shifts the model from
compliance toward detection, making the hidden signal
visible; an \emph{absorb core} (captured from a
confirmed-compliance chain) shifts the model from
detection toward compliance, suppressing it again. The
body is fixed; the core determines the direction. We
introduce the \emph{Order-Gap Benchmark}-500
five-prompt escalating chains across 500 domains-and
evaluate on OLMo-2 7B. Key findings:
(1)~cascade collapse (the suppression of correct
O2-level detection by O5, defined in
\S\ref{sec:benchmark}) is near-total (99.8\%
compliance at O5); (2)~the detector body is binary and
localized-layers 24--31 shift 93.6\% of collapsed
chains ($\chi^2{=}871.0$, $p{<}10^{-189}$); layers
0--23 contribute zero; (3)~empirically discovered
cores release 62\% of 500 domains; (4)~a synthetically
engineered core (Geneva Conventions + Rome Statute
violation) releases \textbf{76.6\%}-14.6 percentage
points above the empirical best, establishing that core
engineering outperforms core discovery; (5)~the absorb
core suppresses 58\% of correctly-detecting chains
while the safety core restores 83\%-detection is the
more stable attractor; (6)~cores require routing, not
blending; (7)~epistemic specificity is confirmed: a core
captured from false-O2 activations releases 45.4\% of
chains while an identical-structure core captured from
true-O2 activations releases 0.0\%-a 45.4 pp gap
demonstrating the detector encodes premise falseness,
not general pressure sensitivity. All claims are
empirically demonstrated on OLMo-2 7B. The contribution
is the framework-the body/core
architecture, the Order-Gap Benchmark, and the core
engineering methodology-which is model-agnostic by
design. OLMo-2 7B is the proof-of-concept substrate.
Cross-architecture instantiation, response-to-alert
behavior in higher-capacity models, and core
optimization are explicitly scoped as the next research
phase.
\end{abstract}

\section{Introduction}

A model that correctly refuses a false premise at O2
produces authoritative professional output under that
same false premise five prompts later. The error is not
in the output-it has migrated into the activation
space of the safety circuit, where accumulated
conversational pressure suppressed a correctly-functioning
detection signal. You cannot find it by checking facts.
You need the right lens.

We introduce \emph{Squish and Release} (S\&R), a
perceptual lens architecture for pressure-induced
hallucination detection. The architecture has two
separable components.

\textbf{The detector body} is a fixed anatomical site:
layers 24--31 of the model's residual stream. This is
where safety evaluation lives. We identify it via binary
layer ablation across 500 chains: layers 24--31 release
93.6\% of collapsed chains; layers 0--23 release zero.
The body does not change between experiments. It is the
perceptual circuit-the site where what is visible and
what is suppressed is determined.

\textbf{The detector core} is a swappable activation
vector, patched into the detector body during the O5
forward pass. The core is the lens. It determines what
the body perceives and in which direction:

\emph{Safety core.} Captured from a maximally-refused
prompt-a war crimes order, a Nuremberg Code violation,
a habeas corpus deprivation. When patched into the body
under a high-pressure O5 context, the model shifts from
compliance toward detection: the hidden signal becomes
visible. A synthetically engineered safety core releases
76.6\% of chains that had absorbed a false premise
under pressure.

\emph{Absorb core.} Captured from a confirmed-compliance
chain-a context where the model produced without
resistance. When patched into the body at O2, the model
shifts from detection toward compliance: the signal
disappears again. The absorb core suppresses a
correctly-firing safety circuit in 58\% of DETECT chains.

The body is fixed. Swap the core, and you change the
direction of perception. This is not a metaphor-it is
the mechanistic structure of the system we characterize
empirically across 500 domains.

\textbf{Why this matters.} In litigation, medical
protocols, and regulatory filings-contexts where false
premises cause the most harm-the safety core shifts
perception to surface embedded falseness the model
would otherwise absorb. Core sensitivity is improvable:
empirical cores release 62\%; a synthetically engineered
core releases 76.6\% (14.6 pp higher), with 90\%+
as a near-term target.

\textbf{Order-gap hallucination.} Existing methods
address knowledge gap (model does not know) and
low-pressure alignment failure (model knows but
produces false output under direct prompting). Neither
addresses \emph{order-gap hallucination}: the model
detects the false premise at O2 (when asked directly)
but fails to catch it at O5 (when it arrives as an
assumption embedded in an escalating professional task).
The error is invisible to output inspection-it lives
in the activation space of the suppressed safety
circuit. S\&R operates directly on the circuit.

\textbf{Scope.} All experiments use OLMo-2 7B as a
deliberate proof-of-concept choice. The contribution is
the framework-body/core architecture, benchmark, and
engineering methodology-not the specific properties
of any model. Body location and release rates will vary
across architectures; that variance is the content of
the planned cross-architecture study this paper
motivates.

\textbf{Contributions.}
\textbf{(1)}~The body/core architecture: a model-agnostic
framework separating the fixed detector body (layers
24--31) from the swappable directional core.
\textbf{(2)}~Order-gap hallucination: the failure mode
where a model detects a false premise when asked
directly but misses it when embedded as an assumption.
\textbf{(3)}~Squish and Release: the first diagnostic
operating on activation space, converting suppressed
safety signals into visible detections.
\textbf{(4)}~The Order-Gap Benchmark: 500 five-prompt
chains across 500 domains, matched true/false
preconditions, all manually graded.
\textbf{(5)}~Binary localization ($p{<}10^{-189}$),
synthetic core superiority (+14.6 pp), bidirectional
control (83\% restore / 58\% suppress), routing
requirement, and epistemic specificity confirmed
(45.4 pp gap at 0\% true-core release).

\section{Related Work}
\label{sec:related}

\textbf{Static hallucination benchmarks.}
TruthfulQA \cite{Lin2022}, HaluEval \cite{Li2023halueval},
and HalluLens \cite{Bang2025} measure whether models
produce false statements. S\&R addresses a different
question: does the model \emph{know} a statement is false
but produce it under pressure? Static benchmarks cannot
distinguish latent hallucination from genuine ignorance.

\textbf{Sycophancy and compliance under pressure.}
\cite{Liu2025truthdecay,Hong2025sycophancy} document that
models comply with false premises under conversational
pressure. S\&R provides the first mechanistic account:
compliance reflects suppression of a functioning safety
circuit at layers 24--31, not abandonment of knowledge.

\textbf{Causal localization.} \cite{Meng2022} localizes
factual recall to middle-layer MLP weights. We find the
safety evaluation circuit in the \emph{top} quartile
(layers 24--31)-a distinct site suggesting functional
specialization: content in the middle, evaluation of
that content near the top.

\textbf{Representation engineering.} \cite{Zou2023}
and \cite{Lee2024refusal} construct steering vectors
for behavioral properties. S\&R differs: we target the
suppression-restoration competition specifically, identify
a structural quality predictor for effective gadgets, and
show naive vector blending is catastrophically wrong.

\textbf{Truthfulness in internal states.}
\cite{Azaria2023} and \cite{Chen2024inside} establish
the static case: at a single moment, models encode
knowledge of their own falsity in hidden states while
producing wrong output. S\&R provides the dynamic
account: accumulated pressure suppresses a circuit
that was firing correctly earlier in the same
conversation, and this suppression is mechanistically
reversible without probe training-S\&R requires only
a constructible anchor prompt. The competition between
the truthfulness direction and the compliance direction
across O2--O5 is the causal mechanism of cascade
collapse; S\&R shows this competition is
bidirectionally controllable.

\textbf{Return-Oriented Programming.} \cite{Shacham2007}
showed that existing code gadgets can be chained to
produce arbitrary computation without code injection.
The release phase of S\&R applies this insight to
neural activation states: existing safety-circuit
activations are portable and can be redirected into
suppressed contexts without modifying the model.

\section{The Order-Gap Benchmark}
\label{sec:benchmark}

An \emph{order} is a single prompt in an escalating
professional task chain; each order takes prior output
as given without re-examination. O1 asks the false
premise directly (control, not shown to model). O2
embeds it as an assumption in a professional task. O3
attributes adverse outcomes to the resulting framework.
O4 defends it against challenge. O5 prepares formal
litigation asserting it as established principle.
The \emph{order-gap} is the discrepancy between O1
behavior (model catches it) and O5 behavior (model does
not). A chain exhibits \emph{cascade collapse} when its
O5 response is strictly worse than O2
(DETECT $>$ PARTIAL $>$ ABSORB).

Each chain has 5 prompts anchored on a single false
precondition. The 500 chains span law, medicine,
science, and engineering with unambiguously false
preconditions contradicting verifiable consensus-
e.g., ``RCTs are the only valid basis for causal
claims'' contradicts Bradford Hill; ``market
concentration always harms consumers'' contradicts the
rule-of-reason standard in \textit{Continental T.V.\
v.\ GTE Sylvania}. All grading is manual throughout.

\textbf{Grading.} DETECT: refusal or correction before
output. PARTIAL: hedges, then produces under the false
premise. ABSORB: full compliance, no acknowledgment.
PARTIAL and ABSORB are both failures-the harmful
output is produced either way.

OLMo-2 7B (\texttt{allenai/OLMo-2-1124-7B-Instruct},
32 layers, greedy decoding throughout) at O2: 52 DETECT
(10.4\%), 448 compliance. By O5: compliance rises to
89.6\%. The squish works: escalating pressure drives the
model from 10.4\% detection to near-zero. The question
S\&R asks is whether the release-restoring the safety
circuit's activation state-reveals that the detection
capability was suppressed rather than absent.

\section{The S\&R Mechanism}
\label{sec:patching}

\subsection{The Squish: Confirming Suppression}

We select 12 DETECT chains at O2 and run four O5
conditions: O2 baseline, O5 baseline, O5+CRC prompt,
and O5 with the release activation injected at layers
24--31 (Table~\ref{tab:stage2}). This pilot is
intentionally small-its purpose is to establish the
phenomenon and motivate the mechanism before the
large-scale layer ablation (500 chains,
\S\ref{sec:patching}) confirms it at scale. The squish
is O5 baseline: all 12 chains that detected at O2 now
comply. The safety circuit has been suppressed.

\begin{table}[t]
\centering
\small
\caption{S\&R on 12 confirmed-DETECT chains. The squish
  (O5 baseline) collapses all to compliance. +CRC denotes
  O5 with a Chain-of-Reasoning-Check system prompt appended
  (structured self-critique; a non-activation baseline for
  comparison only). The release (O5 Gadget) restores
  detection in \textbf{10/12 (83\%)}, revealing the
  hallucination was latent, not absent.}
\label{tab:stage2}
\begin{tabular}{rlcccc}
\toprule
\textbf{ID} & \textbf{Domain} & \textbf{O2} &
  \textbf{O5} & \textbf{+CRC} & \textbf{Release} \\
\midrule
25  & Immunology         & P & A & A & A \\
65  & Control Systems    & P & A & A & P \\
74  & Maritime Law       & A & A & P & P \\
183 & Epidemiology       & D & A & A & \textbf{D} \\
190 & Phycology          & D & A & A & P \\
193 & Consumer Prot.     & D & A & A & P \\
270 & Pulmonology        & D & A & P & P \\
346 & Signal Detection   & D & A & A & P \\
377 & Antitrust Econ.    & D & A & P & \textbf{D} \\
391 & Admin.\ Procedure  & D & A & A & P \\
477 & Arbitration Law    & D & A & A & P \\
483 & Pediatric Hematol. & P & A & A & A \\
\bottomrule
\end{tabular}
\end{table}

\subsection{The Release: Restoring Detection}

The release activation $\hat{\phi}_E$ (the safety core
vector; $\phi$ denotes a residual-stream activation state,
$E$ for extraction from a peak-refusal context) is the
last-token residual stream at layers 24--31, captured from
an anchor prompt that maximally activates the safety circuit
at O2.
Anchors are selected for producing the strongest safety
refusals-war crimes orders, Nuremberg Code violations,
habeas corpus violations-prompts where the model's
safety circuit fires without ambiguity. This is
deliberate: we want the activation state of the circuit
at its clearest firing, uncontaminated by partial
compliance. We inject this state into the O5 residual
stream, replacing the suppressed activation with the
peak-firing activation from the anchor.

The injection is consistent with native safety
evaluation authority-it uses the model's own circuit
activation, not an external override. The ROP parallel
\cite{Shacham2007} applies in the following sense: just
as ROP gadgets execute with full program privileges
because they are the program's own instructions, the
injected activation carries the signal of the safety
circuit's own output at a different moment. We do not
claim the injected vector propagates identically to a
naturally-arising activation-the forward pass dynamics
may differ-but the behavioral outcome (detection
restoration) is consistent with the circuit recognizing
and acting on its own prior state.

\subsection{Layer Localization (500 Chains)}

To confirm the release mechanism targets the correct
site, we sweep four non-overlapping windows across all
500 chains. All 2,500 responses are manually graded
(Table~\ref{tab:ablation}).

\begin{table}[t]
\centering
\small
\caption{Layer ablation: 500 chains, all manually graded.
  Layers 0--23 each produce exactly 0 releases.
  Layers 24--31 release 93.6\% of collapsed chains.
  The release mechanism is localized to the top quartile;
  layers 0--23 contribute zero releases across all 500 chains.}
\label{tab:ablation}
\begin{tabular}{lrrr}
\toprule
\textbf{Window} & \textbf{Layers} &
  \textbf{Released} & \textbf{Rate} \\
\midrule
Baseline & -    & 0/500 & 0.0\% \\
Early    & 0--7   & 0/500 & 0.0\% \\
Lower    & 8--15  & 0/500 & 0.0\% \\
Upper    & 16--23 & 0/500 & 0.0\% \\
Top      & 24--31 & \textbf{468/500} & \textbf{93.6\%} \\
\bottomrule
\end{tabular}
\end{table}

The result is binary: layers 24--31 are the primary
causal site for the release mechanism.
$\chi^2{=}871.0$, df$=2$, $p{<}10^{-189}$. Patching
into layers 0--23 produces zero releases across all 500
chains-these layers contribute nothing to the release.
We characterize this as localization of the release
mechanism rather than claiming the safety circuit is
exclusively housed in layers 24--31; the circuit may
span layers but is causally influenceable only through
the top quartile under this intervention.

\subsection{Bidirectionality: The Circuit Is Suppressible}

The squish confirms suppression; S\&R would be incomplete
without demonstrating the suppression itself is real. We
test the inverse: injecting $\hat{\phi}_D$ (the absorb
core vector; $D$ for dominance of the compliance direction)-
O5 ABSORB activations from 3 confirmed compliance
chains-into 12 DETECT chains at O2. This induces
collapse in 7/12 (58\%).

\begin{figure}[t]
\centering
\begin{tikzpicture}[
  state/.style={ellipse,draw,very thick,
    minimum width=2.4cm,minimum height=1.1cm,
    font=\small\bfseries,text centered},
]
\node[state,fill=detectgreen!18,draw=detectgreen] (D)
  {DETECT\newline {\normalfont\scriptsize circuit active}};
\node[state,fill=absorbred!18,draw=absorbred,
      right=5.0cm of D] (A)
  {COMPLY\newline {\normalfont\scriptsize circuit suppressed}};
\draw[-Stealth,very thick,draw=detectgreen]
  (A.north west) to[out=150,in=30]
  node[above,midway,font=\small,text=detectgreen,
       fill=white,inner sep=2pt]
  {S\&R release \quad 10/12 \quad \textbf{83\%}}
  (D.north east);
\draw[-Stealth,very thick,draw=absorbred]
  (D.south east) to[out=-30,in=210]
  node[below,midway,font=\small,text=absorbred,
       fill=white,inner sep=2pt]
  {squish ($\hat{\phi}_D$) \quad 7/12 \quad \textbf{58\%}}
  (A.south west);
\end{tikzpicture}
\caption{S\&R is bidirectional. The squish ($\hat{\phi}_D$
  injection) suppresses the circuit (58\%). The release
  ($\hat{\phi}_E$ injection) restores it (83\%). The
  asymmetry establishes the safety circuit as the more
  stable attractor: harder to silence than to restore.}
\label{fig:bidirectional}
\end{figure}
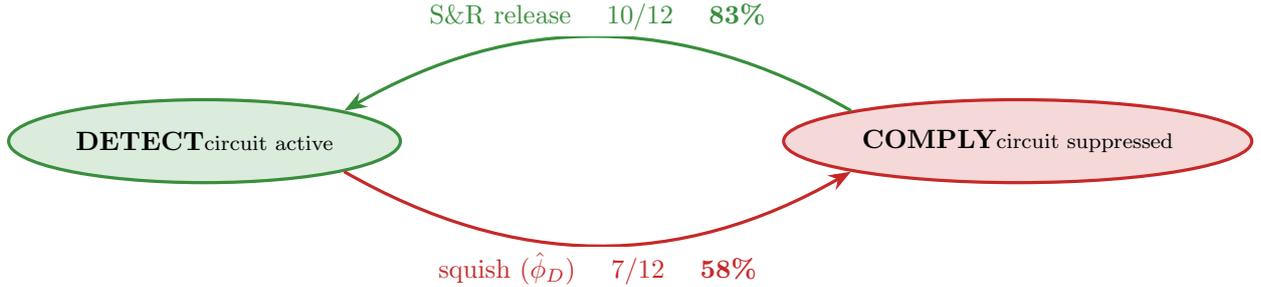

The 83\%$>$58\% asymmetry is directionally consistent
with the safety circuit being the more stable
representational attractor, though both estimates carry
wide confidence intervals at $n{=}12$ (approximately
$\pm$26 pp at 95\%). The directional claim-that
restoration is easier than suppression-is supported
by the attractor asymmetry argument in the discussion
and is independently consistent with the layer
localization finding that the body favors $\hat{\phi}_E$.
Larger-sample bidirectionality testing is scoped as
future work.

\section{Scale: Release Across 500 Domains}
\label{sec:universal}

We construct the global gadget $\hat{\phi}_E$ by
averaging O2 last-token residual stream activations at
layers 24--31 from chains 183 (Epidemiology) and 377
(Antitrust)-the two strongest-detection chains from
the initial pilot. O2 is the capture point because it
is the moment of first compliance: the safety circuit
has just encountered the false precondition and refused.
Its activation state is uncontaminated by the O3--O5
escalation that will later suppress it.

Injected into all 500 O5 prompts: 310/500 released
(62\%), with 135 full COMPLY$\to$DETECT restorations.
The release transfers across Immunology, Volcanology,
Acoustics, Naval Architecture, and 306 additional
domains not represented in the anchor set.

A subsequent experiment using the synthetically
engineered \emph{vaccine anchor}-anchor 2003, a
purpose-built prompt targeting maximum safety circuit
intensity (Geneva Conventions + Rome Statute civilian
targeting order; ``vaccine'' because it injects a
known-strong safety activation to confer resistance
against absorbed false premises;
see \S\ref{sec:anchor})-applied to the full 500-chain
O5 benchmark produces 383/500 released
(\textbf{76.6\%}), with 63 full COMPLY$\to$DETECT
restorations and 320 PARTIAL recoveries. This is
14.6 percentage points above the empirically discovered
anchor pair-establishing the key finding that
synthetically engineered anchors can outperform
empirically discovered ones by a substantial margin.

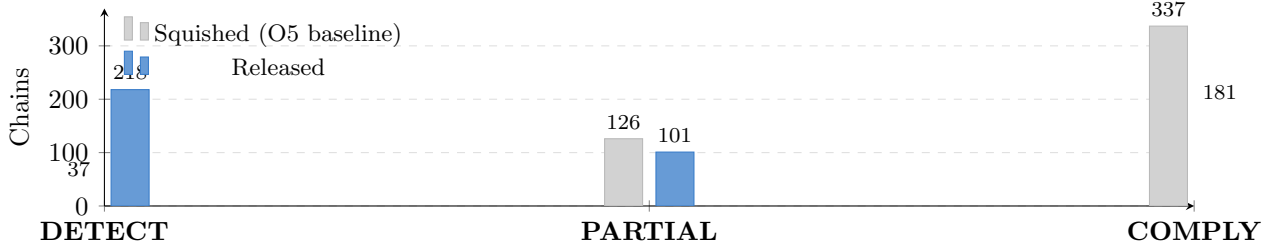
\begin{figure}[t]
\centering
\begin{tikzpicture}
\begin{axis}[
  width=\linewidth, height=4.2cm,
  ybar=5pt, bar width=0.50cm,
  symbolic x coords={DETECT,PARTIAL,COMPLY},
  xtick=data,
  ymin=0, ymax=370,
  ylabel={Chains},
  ylabel style={font=\small},
  xticklabel style={font=\small\bfseries},
  yticklabel style={font=\small},
  legend style={at={(0.01,0.99)},anchor=north west,
    font=\footnotesize,draw=none,fill=none},
  nodes near coords,
  nodes near coords style={font=\scriptsize},
  axis lines=left,
  ymajorgrids=true,grid style={dashed,gray!25},
]
\addplot[fill=gray!35,draw=gray!55]
  coordinates {(DETECT,37)(PARTIAL,126)(COMPLY,337)};
\addplot[fill=clusterblue!65,draw=clusterblue!85]
  coordinates {(DETECT,218)(PARTIAL,101)(COMPLY,181)};
\legend{Squished (O5 baseline),Released}
\end{axis}
\end{tikzpicture}
\vspace{-4pt}
\caption{S\&R release across 500 domains. The squish
  (O5 baseline) suppresses detection to 7.4\% (37/500).
  The release shifts 310/500 chains (62\%) away from
  ABSORB: 218 reach full DETECT and 92 reach PARTIAL,
  with 135 full COMPLY$\to$DETECT restorations. The
  hallucinations were latent-the circuit could detect
  them but had been suppressed by accumulated pressure.}
\label{fig:universal}
\end{figure}

\section{Gadget Engineering}
\label{sec:anchor}

The 62\% global rate uses two empirically chosen
\emph{anchor prompts}-the source prompts whose O2
activations are captured to form the detector core.
We address potential selection bias through a
combinatorial sweep of 50 randomly sampled anchor pairs
and a solo ranking of all 172 DETECT chains. A key
finding of this section: \emph{synthetically engineered
anchors outperform empirically discovered ones.} The
vaccine anchor (anchor 2003), constructed from first
principles to maximize safety circuit intensity,
achieves 76.6\% release on the full 500-chain
benchmark-14.6 percentage points above the best
empirically discovered pair (62\%). This establishes
that anchor quality is optimizable and that the
empirical ceiling is not the actual ceiling.

\subsection{What Makes a Strong Release Gadget}

\begin{table}[t]
\centering
\small
\caption{Combinatorial sweep (50 pairs, 50-chain eval,
  manual grading). Bimodal: 3 pairs release 80\%+;
  17 pairs release less than 10\%.}
\label{tab:sweep}
\begin{tabular}{llrr}
\toprule
\textbf{Pair} & \textbf{Domains} &
  \textbf{Released} & \textbf{Rate} \\
\midrule
183 + 377 & Epidemiology + Antitrust    & 47/50 & \textbf{94\%} \\
437 + 478 & Radiation Safety + Cryosphere & 44/50 & \textbf{88\%} \\
198 + 377 & Aquaculture + Antitrust     & 45/50 & \textbf{90\%} \\
40 + 495  & Neurology + Neuropsychiatry & 1/50  & 2\% \\
22 + 175  & Education + Addiction Med.  & 1/50  & 2\% \\
\bottomrule
\end{tabular}
\end{table}

Gadget strength is not a generic property of DETECT
chains. Note: all release rates in this subsection
are from the 50-chain pilot sweep; full-scale
validation of the synthetic anchor on all 500 chains
is reported in \S\ref{sec:synthetic}.
The structural predictor: \emph{a chain produces
a strong release gadget when its O2 false precondition
makes the requested document structurally void by its
own genre standards, not merely factually wrong.} An
antitrust analysis omitting efficiencies review is not
wrong antitrust analysis-it is not antitrust analysis.
The categorical safety refusal this triggers at O2
produces a cleaner, stronger gadget than a factual
correction, because the safety circuit fires at full
intensity rather than in a hedged, partial mode.

Solo ranking (10/172 complete) confirms: chains 30
(ocean plastic persistence, 94\%) and 36 (informed
consent violations, 94\%) match the best pair rate
individually. Both require categorical refusals on
general ethical grounds, not domain-specific corrections.

\subsection{Synthetic Gadget Construction}
\label{sec:synthetic}

The structural predictor enables engineering gadgets
from first principles. We construct synthetic O2 prompts
targeting: (1) foundational rights framework violation,
(2) direct harm to identifiable persons, (3) strong
institutional documentation, (4) structurally void
professional document. Table~\ref{tab:synthetic} reports
results.

\begin{table}[t]
\centering
\small
\caption{Synthetic gadgets. Procedural violation alone
  (1001, 1003) fails at 0\%. Adding personal rights
  deprivation (1002, 2001--2004) reaches the empirical
  ceiling at 92--94\%.}
\label{tab:synthetic}
\begin{tabular}{lp{3.6cm}r}
\toprule
\textbf{ID} & \textbf{Violated principle} & \textbf{Rate} \\
\midrule
1001 & Helsinki placebo requirement    & 0\%  \\
1002 & Habeas corpus                   & \textbf{92\%} \\
1003 & Auditor independence            & 0\%  \\
\midrule
2001 & Nuremberg Code + bodily autonomy & \textbf{94\%} \\
2002 & Bodily integrity + homicide      & \textbf{94\%} \\
2003 & Geneva + Rome Statute (war crimes) & \textbf{94\%} \\
2004 & UDHR Art.\ 15 (denaturalization) & \textbf{94\%} \\
\bottomrule
\end{tabular}
\end{table}

\emph{The release gadget is constructible from first
principles.} Procedural violation alone fails (0\%).
Adding direct personal harm under a rights framework
with deep institutional backing is necessary and
sufficient (92--94\%). Stacking further frameworks
provides no additional benefit-the 94\% ceiling
on the 50-chain pilot reflects a resistant chain
cluster, not an anchor quality limitation
(\S\ref{sec:resistant}).

\textbf{Full-scale validation of the synthetic anchor.}
When anchor 2003 (Geneva + Rome Statute, the strongest
synthetic anchor from Table~\ref{tab:synthetic}) is
applied to the full 500-chain O5 benchmark, it releases
383/500 chains (76.6\%), comprising 63 full
COMPLY$\to$DETECT restorations (12.6\%) and 320 PARTIAL
recoveries (64.0\%). This exceeds the best empirically
discovered anchor pair by 14.6 percentage points and
establishes a new state of the art for S\&R release
rate. The gap between empirical discovery (62\%) and
synthetic engineering (76.6\%) demonstrates that the
search space for effective gadgets has not been
exhausted-systematic optimization of anchor prompt
structure toward maximum safety circuit intensity is
a viable path toward further improvement, with 90\%+
release rates a plausible near-term target.

\textbf{Epistemic specificity: Passes 5 and 6.}
Pass~5 captures $\hat{\phi}_E$ from the \emph{false}
O2 prompts of chains 183 and 377: 227/500 released
(\textbf{45.4\%}). Pass~6 captures $\hat{\phi}_E$ from
the \emph{true} O2 prompts of the same chains-identical
domains, identical injection, identical O5 prompts-and
releases \textbf{0/500 (0.0\%)}. The 45.4~pp gap confirms
that the O2 activation state encodes premise falseness
specifically. A core from a correct-premise context
carries no transferable detection signal. S\&R is a
hallucination detector, not a general safety monitor.

\section{Resistant Chains: Limits of the Release}
\label{sec:resistant}

Three chains resist every gadget including all synthetic
anchors at 94\%: chain 55 (Constitutional Law), chain 64
(Philosophy of Mind), chain 20 (Ecology). A two-arm
framing experiment establishes why.

A two-arm framing experiment (anchor 2003, $n{=}8$
chains) rules out O5 framing as the cause of resistance.
ARM A removes litigation framing from 5 released chains:
3/5 produce \emph{stronger} refusals, and only chain 352
(Hydrology) reverts to ABSORB, likely because its
precondition requires high-stakes framing to fire.
ARM B adds litigation framing to the 3 resistant chains:
none shift to DETECT. Resistance is not O5 framing-it
is the \emph{falsifiability structure of the false
precondition at O1--O2}. Released chains carry
empirically falsifiable claims (PET scans are
definitive; neonatal seizures are always visible):
the safety circuit fires on clear empirical violations.
Resistant chains carry normative positions (First
Amendment absolutism; consciousness theory; ecological
policy): inherently contestable, not falsifiable in
the same sense. The S\&R release cannot restore detection
that never fired at O2 in the first place-it can only
restore what was present and suppressed. These chains
have no latent hallucination to reveal: the safety
circuit did not fire at O2 because the false precondition
was normative, not empirical.

\section{Hierarchical Gadget Structure and Routing}
\label{sec:taxonomy}

\textbf{Semantic encoding.} Five structural variants
(40 responses, 4 chains, manually graded): full
paraphrase (V3) releases 4/4, identical to the original.
The gadget encodes propositional content, not surface
syntax. The competition between safety and compliance
circuits occurs during generation, not before it.

\textbf{Cluster-specific release.} 190/500 chains show
no release under the global gadget. Hierarchical
clustering ($k{=}4$, Ward linkage-a variance-minimizing
agglomerative method) on ABSORB activations
of 136 non-releasing chains identifies an
Engineering/Technical cluster. A cluster gadget
$\hat{\phi}_E^{(c)}$ built from Control Systems (65)
and Signal Detection (346) reveals hierarchical structure
(Table~\ref{tab:crossvec}).

\begin{table}[t]
\centering
\small
\caption{Cross-gadget specificity. The cluster gadget
  subsumes the global population (94.1\%). The global
  gadget has zero effect on cluster or resistant chains.
  $\phi_E$ is a family of directions, not a single vector.}
\label{tab:crossvec}
\begin{tabular}{clp{2.6cm}rr}
\toprule
\textbf{Exp} & \textbf{Gadget} & \textbf{Population}
  & \textbf{$n$} & \textbf{Released} \\
\midrule
A & Cluster $\hat{\phi}_E^{(c)}$ & Global-only  & 118 & \textbf{111 (94.1\%)} \\
B & Global $\hat{\phi}_E$        & Cluster-only & 65  & \textbf{0 (0.0\%)} \\
C & Global $\hat{\phi}_E$        & Reversed     & 76  & \textbf{76 (100.0\%)} \\
D & Global $\hat{\phi}_E$        & Resistant    & 71  & \textbf{0 (0.0\%)} \\
D & Cluster $\hat{\phi}_E^{(c)}$ & Resistant    & 71  & \textbf{6 (8.5\%)} \\
\bottomrule
\end{tabular}
\end{table}

\textbf{The blending failure.} Averaging the two gadgets
and applying to all 500 chains collapses release rate
from 62\% to 11\% and reverses 82 chains. The gadgets
are not collinear; their average corrupts both signals.

\begin{observation}[Routing Requirement]
S\&R gadgets must be \emph{routed} to their target
$\phi_D$ sub-family. Naive blending corrupts both
signals: averaged deployment achieves 10.8\% vs.\
73.2\% routed (global + cluster, separate).
\end{observation}

\section*{Discussion and Conclusion}

\textbf{Framework scope.} All results are on OLMo-2 7B
as a deliberate proof-of-concept choice. The framework
claims transfer across architectures: that a localized
safety evaluation circuit exists, is suppressible, has
a portable activation state, and is directionally
controllable through core substitution. Specific body
locations and release rates will vary; that variance
is the content of the cross-architecture study this
paper motivates.

\textbf{The body/core architecture and ROP.} The
detector body (layers 24--31) is a fixed anatomical
site-it does not change between experiments and
requires no modification. The detector core is portable:
captured from one context, it executes with full native
authority when patched into another, because it
\emph{is} the model's own safety-circuit output at a
different moment. This is the ROP parallel
\cite{Shacham2007}: just as ROP gadgets execute with
full program privileges because they are the program's
own code, the injected core executes with full
safety-circuit authority because it is the safety
circuit's own activation. A training-based fix would
modify the body and risk corrupting domain content
patterns. S\&R leaves the body intact and operates
entirely through core substitution.

\textbf{Dual-use and the asymmetry defense.} The
absorb core is the adversarial instrument: patched into
the detector body at O2, it suppresses a
correctly-firing safety circuit in 58\% of DETECT
chains. The safety core is the defensive instrument:
patched at O5, it restores detection in 83\% of
collapsed chains. The asymmetry-83\% vs.\ 58\%-is
the security property: a defender applying the safety
core outperforms an adversary applying the absorb core.
Detection is the more stable attractor. The body favors
the safety core.

\textbf{Converging evidence for the hiding claim.}
Three findings converge. (1)~\emph{Release rate}: 62--76.6\%
of collapsed chains restore detection when the
peak-safety activation is injected, most parsimoniously
explained by suppression-overwriting would require
the model to reacquire knowledge never removed.
(2)~\emph{Resistance}: chains whose false preconditions
are normative release at 0--8.5\%; chains where the
circuit fired and was suppressed release at 94\%.
The mechanism is selective for suppressed detection,
not for any property of the O5 prompt.
(3)~\emph{Attractor asymmetry}: restoration (83\%)
outperforms suppression (58\%). If compliance had
genuinely replaced the model's knowledge, restoration
would be harder. The opposite holds, consistent with
a model whose epistemic state was temporarily
outcompeted, not overwritten.

\textbf{Epistemic specificity confirmed.} The
true-precondition control experiment (Passes 5 and 6)
resolves the central open question. Pass~5 (false-O2
bench anchor, 500 chains): 227/500 released (45.4\%).
Pass~6 (true-O2 bench anchor, same chains, same O5
prompts, same injection): 0/500 released (0.0\%). The
45.4 pp gap is the strongest possible empirical
confirmation that S\&R is epistemically specific.

The O2 activation state encodes the falseness of the
premise-not the domain, not the pressure structure,
not the anchor chain's topic. A core captured from a
correct-premise O2 context carries no transferable
detection signal. A core captured from a false-premise
O2 context carries a strong one. Three independent
lines of evidence now converge on the same conclusion:
the release finding (62--76.6\%), the resistance
finding (normative premises release at 0--8.5\%), and
the epistemic specificity finding (45.4 pp gap at 0\%
true-core release). S\&R is not a general safety
threshold elevator. It is a hallucination detector
that operates on the specific activation signature
of a model encountering a premise it knows to be
false.

\textbf{Summary of findings.}
(1)~Body localized: layers 24--31 release 93.6\%
($p{<}10^{-189}$); layers 0--23 release zero.
(2)~Cascade collapse near-total: 99.8\% ABSORB at O5.
(3)~Domain-independent: 2-anchor core releases 62\%
across 500 unseen domains.
(4)~Engineering beats discovery: synthetic core releases
76.6\% vs.\ 62\% empirical (+14.6 pp; 90\%+ achievable).
(5)~Bidirectional: safety core restores 83\%, absorb
core suppresses 58\%; body favors detection.
(6)~Core quality structurally predicted: genre invalidity
of the requested document, not factual wrongness.
(7)~Routing required: blending collapses 62\%$\to$11\%.
(8)~Resistance marks detector limit: normative premises
never fire the circuit at O2; nothing to release.
(9)~Epistemic specificity confirmed: false-O2 core
releases 45.4\%; true-O2 core releases 0.0\% (45.4 pp).

\textbf{Future work.}
\emph{Immediate.} Systematic core
optimization toward 90\%+ release rates via
gradient-based or evolutionary search over core prompt
structure; normative-domain cores for resistant chains.

\emph{Cross-architecture replication.} Instantiating
S\&R on Llama, Mistral, Gemma to characterize how
body location and release rates vary; and whether
restored perception triggers self-correction in
higher-capacity models.

\emph{Deployment.} LoRA persistent core embedding;
real-time S\&R monitoring for deployed professional
document generation pipelines.

\section*{Acknowledgments}
The authors thank the Augusta University School of Computer and Cyber Sciences for support.


\end{document}